\documentclass{article}

\usepackage[preprint]{neurips_2026}
\usepackage{multirow}
\usepackage{graphicx}
\usepackage{pifont}
\usepackage[most]{tcolorbox}

\usepackage[utf8]{inputenc} % allow utf-8 input
\usepackage[T1]{fontenc}    % use 8-bit T1 fonts
\usepackage{hyperref}       % hyperlinks
\usepackage{url}            % simple URL typesetting
\usepackage{booktabs}       % professional-quality tables
\usepackage{amsfonts}       % blackboard math symbols
\usepackage{nicefrac}       % compact symbols for 1/2, etc.
\usepackage{microtype}      % microtypography
\usepackage{xcolor}         % colors
\usepackage{graphicx}

\usepackage{amssymb}  % newly added
\usepackage{amsmath}  % newly added
\usepackage{makecell} % newly added
\usepackage{fontawesome5}
\setcitestyle{numbers,square}

\title{Native Audio-Visual Alignment for Generation}

\author{%
\textbf{Longbin Ji}$^{*}$ \quad
\textbf{Guan Wang}$^{*}$ \quad
\textbf{Xuan Wei} \quad
\textbf{Chenye Yang}  \quad
\textbf{Xiangrui Liu} \\
\textbf{Zhenyu Zhang}$^{\dagger}$ \quad
\textbf{Shuohuan Wang} \quad
\textbf{Yu Sun} \quad
\textbf{Jingzhou He} \\
[2mm]
ERNIE Team, Baidu Inc. \\
[1mm]
\texttt{\{jilongbin, wangguan15, zhangzhenyu07\}@baidu.com} \\
[1mm]
{\small $^{*}$Equal contribution. \quad $^{\dagger}$Corresponding author.} \\
[2mm]
{\small
\raisebox{-0.12em}{\includegraphics[height=0.9em]{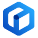}}\,
\textbf{Project page:}
\href{https://ernie-research.github.io/NAVA/}{\texttt{ernie-research.github.io/NAVA}}.
}
}
\begin{document}

\maketitle

\begin{abstract}
Joint audio-video generation aims to synthesize temporally synchronized and semantically coherent visual-acoustic content. 
However, existing open-source methods mainly rely on either dual-tower designs with posterior alignment or fully unified tri-modal designs that mix textual context, audio, and video in one shared space. 
The former weakens fine-grained audio-video co-evolution, while the latter couples semantic conditioning with low-level synchronization.
To address these limitations, we propose \textbf{NAVA}, a \emph{Native Audio-Visual Alignment} framework for joint audio-video generation. 
NAVA is built upon \emph{context-conditioned native audio-visual alignment}: it first establishes audio-video correspondence in a dedicated interaction space, and then uses external context to condition the joint denoising process. 
Specifically, NAVA is instantiated with an \emph{Align-then-Fuse MMDiT} architecture, which transitions from modality-aware audio-video alignment to modality-shared joint denoising. Furthermore, we introduce \emph{Timbre-in-Context Conditioning} to associate reference timbre cues with corresponding speech spans to achieve controllable speech timbre. 
Experiments on Verse-Bench and Seed-TTS, together with a user study, demonstrate that NAVA achieves superior video quality, precise audio-visual synchronization, competitive audio quality, and stronger reference-timbre controllability using only 6.3B parameters.
\end{abstract}

\section{Introduction}

Audio-visual generation has made rapid progress in recent years. Compared with cascaded pipelines that synthesize one modality after another, joint audio-video generation models temporal and semantic correspondences within a unified generation process, thereby reducing error propagation and improving cross-modal coherence.
Although commercial systems such as Seedance~\cite{seedance2026seedance}, Kling~\cite{kling30}, and Veo~\cite{veo31} have demonstrated the potential of joint audio-video synthesis, their architectures and training recipes remain proprietary.
Therefore, recent open-source efforts, including Ovi~\cite{low2025ovitwinbackbonecrossmodal}, LTX~\cite{hacohen2026ltx}, and MoVA~\cite{team2026mova}, have become crucial for reproducible research in audio-visual generation.
% Recent open-source efforts, including Ovi~\cite{low2025ovitwinbackbonecrossmodal}, LTX~\cite{hacohen2026ltx}, and MoVA~\cite{team2026mova}, have therefore become important for reproducible research in audio-visual generation.

\begin{figure*}[t]
    \centering
    \includegraphics[width=\textwidth]{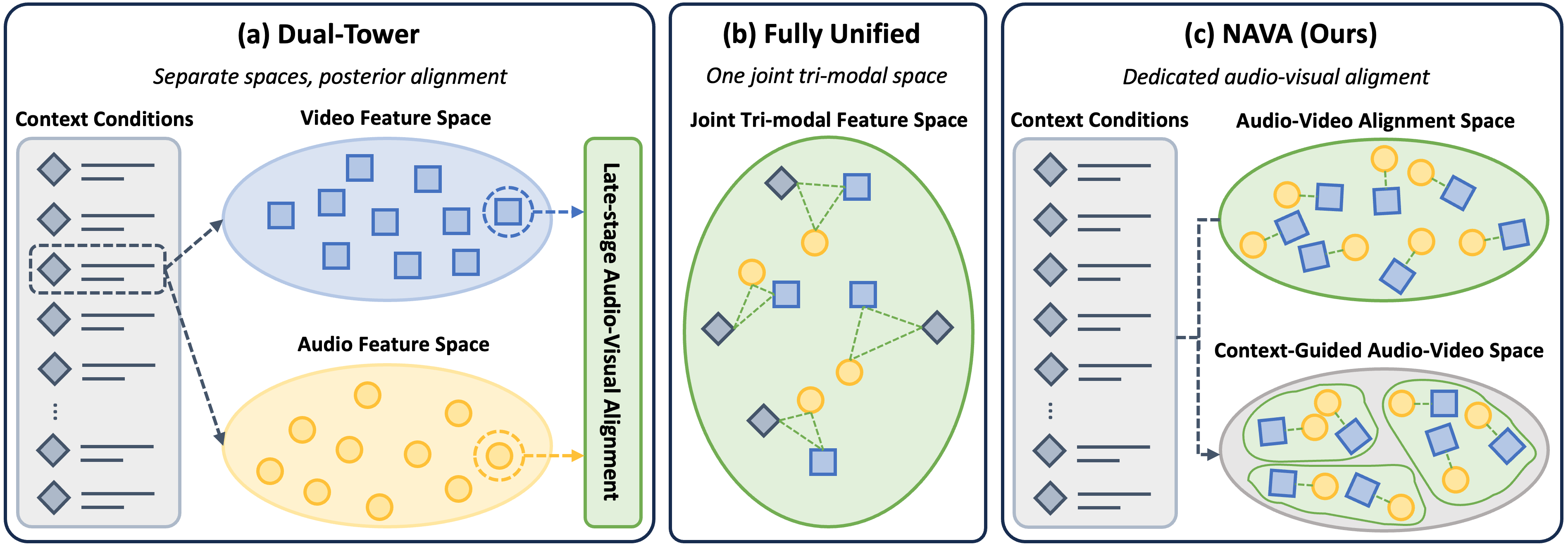}
    \caption{
\textbf{Comparison of different audio-visual generation paradigms.}
(a) \emph{Dual-Tower}: Separate audio and video feature spaces with late-stage cross-modal alignment.
(b) \emph{Fully Unified}: A single tri-modal space that couples context conditioning and synchronization.
(c) \emph{NAVA}: Dedicated audio-video alignment followed by external context conditioning for controllable generation.
}
\label{fig:teaser}
\end{figure*}

Despite this progress, most open-source methods still adopt a dual-tower architecture, where audio and video are generated in separate streams, and cross-modal interaction is introduced through additional alignment modules.
As illustrated in Fig.~\ref{fig:teaser}(a), the paradigm conditions audio and video on textual context in separate feature spaces, and establishes audio-visual correspondence only through late-stage interaction.
However, such posterior alignment weakens the joint evolution of audio and video during generation, making fine-grained synchronization and semantic consistency dependent on auxiliary cross-modal modules rather than a unified generative representation.

More recently, daVinci-MagiHuman~\cite{chern2026speed} moves beyond dual-tower interaction by placing textual context, video, and audio tokens into a unified attention space for end-to-end tri-modal modeling.
As shown in Fig.~\ref{fig:teaser}(b), while the design enables direct tri-modal interaction, it also couples high-level semantic control with low-level audio-visual synchronization.
Consequently, semantic guidance, event correspondence, and temporal alignment are optimized in the same representation space, which may hinder the formation of a dedicated synchronization structure. 
This motivates us to separate audio-video correspondence from context conditioning in a dedicated synchronization space.

In this paper, we propose \textbf{NAVA}, a \emph{Native Audio-Visual Alignment} framework with decoupled context conditioning.
As shown in Fig.~\ref{fig:teaser}(c), NAVA first establishes audio-video correspondence in a dedicated alignment space, and then introduces context as external conditioning to guide the aligned representation.
This formulation differs from both dual-tower methods, which align audio and video only after separate modeling, and fully unified tri-modal methods, which mix context, audio, and video in a shared space.
By decoupling context conditioning from audio-visual synchronization, NAVA focuses its capacity on event-level correspondence, temporal consistency, and collaborative denoising, while remaining compatible with pretrained text-to-video backbones.

To realize this, NAVA employs an \emph{Align-then-Fuse MMDiT} architecture.
It first aligns heterogeneous audio and video representations with modality-aware layers, then applies shared fusion layers for compact collaborative denoising.
Furthermore, we creatively introduce a \emph{Timbre-in-Context Conditioning} mechanism, which treats timbre cues as contextual conditions for specific speech spans, enabling flexible content-timbre binding without auxiliary speaker-control branches.

% Our contributions are summarized as follows
In summary, the main contributions of this paper are as follows:
\begin{itemize}
    \item We propose \textbf{NAVA}, a \emph{Native Audio-Visual Alignment} framework that formulates joint audio-video generation as \emph{context-conditioned native audio-visual alignment}, enabling precise event-level correspondence modeling with pretrained video generation backbones.
    \item We introduce an \emph{Align-then-Fuse MMDiT} architecture for modality-aware audio-video alignment and efficient collaborative denoising, together with \emph{Timbre-in-Context Conditioning} for flexible content-timbre binding across speech segments.
    % \item Extensive experiments and user studies show that \textbf{NAVA} improves audio-visual synchronization, semantic consistency, visual quality, and timbre-controllable generation over representative dual-tower and fully unified baselines under multiple evaluation protocols.
    \item Extensive experiments and user studies demonstrate that \textbf{NAVA} significantly outperforms representative dual-tower and fully unified baselines, achieving superior audio-visual synchronization, semantic consistency, visual quality, and timbre controllability.
\end{itemize}

\section{Method}

\begin{figure*}[t]
    \centering
    \includegraphics[width=\textwidth]{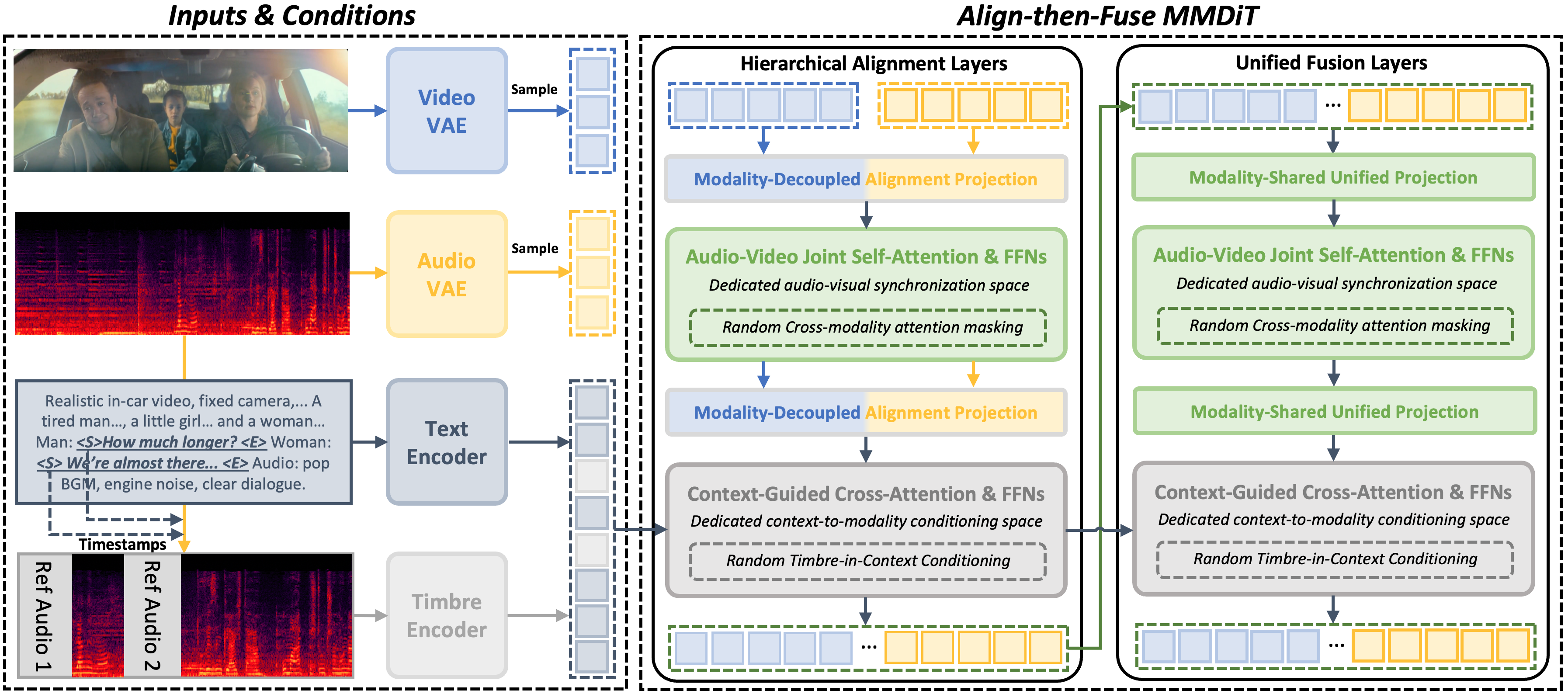}
    \caption{
    \textbf{Overview of NAVA.}
    NAVA adopts an \emph{Align-then-Fuse MMDiT} architecture, which first establishes native audio-video correspondence via \emph{Hierarchical Alignment Layers}, and subsequently performs collaborative denoising using \emph{Unified Fusion Layers}.
    Textual context and optional reference timbre are injected through cross-attention, while \emph{Timbre-in-Context Conditioning} binds timbre cues to speech spans for controllable multi-speaker generation.
    }
    \label{fig:nava_framework}
\end{figure*}

\subsection{Formulation}
\label{sec:formulation}

Let $h_a$, $h_v$, and $c$ denote audio tokens, video tokens, and context tokens, respectively.
The context $c$ mainly contains textual conditions and can be augmented with control signals such as reference timbre embeddings.
We use this notation to abstract how different audio-visual generation paradigms organize audio, video, and context interactions during denoising.

Existing dual-tower methods~\cite{low2025ovitwinbackbonecrossmodal, hacohen2026ltx, team2026mova} maintain separate audio and video generation streams and condition each modality independently:
\begin{equation}
\begin{aligned}
    h_a' &= \mathrm{CrossAttn}(h_a, c), \\
    h_v' &= \mathrm{CrossAttn}(h_v, c).
\end{aligned}
\end{equation}
Audio-visual correspondence is then introduced through additional cross-modal interaction modules:
\begin{equation}
    [\tilde{h}_a, \tilde{h}_v]
    =
    \mathrm{CrossModalAttn}(h_a', h_v').
\end{equation}
This posterior alignment paradigm allows each modality to evolve largely in its own feature space before cross-modal correspondence is explicitly established, making fine-grained synchronization dependent on late-stage interaction.

Fully unified methods~\cite{chern2026speed} instead place context, audio, and video tokens into a single attention space:
\begin{equation}
    [\tilde{h}_a, \tilde{h}_v, \tilde{c}]
    =
    \mathrm{SelfAttn}([h_a, h_v, c]).
\end{equation}
This design enables direct tri-modal interaction, but it also entangles high-level semantic conditioning with low-level audio-video synchronization within the same representation space.

In contrast, \textbf{NAVA} decouples audio-video synchronization from external context conditioning through \textit{context-conditioned native audio-visual alignment}.
Audio and video first interact in a dedicated synchronization space:
\begin{equation}
    [h_a', h_v']
    =
    \mathrm{SelfAttn}([h_a, h_v]),
\end{equation}
where self-attention is applied over the concatenated audio-video token sequence to form event-level correspondences without inserting context as peer tokens.
Context is then injected as external conditioning:
\begin{equation}
    [\tilde{h}_a, \tilde{h}_v]
    =
    \mathrm{CrossAttn}([h_a', h_v'], c).
\end{equation}
In this way, NAVA separates the roles of synchronization and conditioning: joint self-attention learns native audio-video correspondence, while cross-attention provides semantic and controllable guidance from external context.

\subsection{Align-then-Fuse MMDiT}

To instantiate context-conditioned native audio-visual alignment, NAVA adopts an \emph{Align-then-Fuse MMDiT} architecture, as shown in Fig.~\ref{fig:nava_framework}.
Video and audio are first encoded into latent tokens by separate VAEs, while textual context and optional reference-timbre cues are encoded as conditioning tokens.
The architecture follows a progressive design: early layers preserve modality-aware projections to stabilize heterogeneous audio-video interaction, while later layers share generation parameters to encourage compact collaborative denoising.
This yields an align-then-fuse process, where audio and video first establish native correspondence and then evolve jointly in a shared generation space.

\paragraph{Hierarchical Alignment Layers.}
The early layers establish native audio-video correspondence before fully shared generation.
Audio spectrogram latents and video latents differ substantially in spatial-temporal structure, token rate, and feature distribution.
Directly sharing projections from the first layer can therefore force heterogeneous modalities into a common parameterization too early, suppressing modality-specific representations and destabilizing cross-modal interaction.
We address this with \emph{Modality-Decoupled Alignment Projection}, where audio and video tokens are first mapped by modality-specific projections and then placed into a shared audio-video interaction space for stable early-stage correspondence learning.

Within this space, \emph{Audio-Video Joint Self-Attention \& FFNs} perform repeated cross-modal interaction during denoising.
Unlike posterior alignment modules that operate after separate generation streams, this joint interaction allows acoustic patterns and visual dynamics to co-evolve throughout the denoising process.
As a result, event-level correspondences such as speech-lip motion, impact sounds, musical performance, and scene-dependent acoustic changes can be modeled within the generation trajectory itself.
To handle token-rate mismatch, we rescale the rotary positional embedding of audio tokens by
\begin{equation}
    \theta_{\mathrm{rope}} = \frac{TR_v}{TR_a},
\end{equation}
where $TR_v$ and $TR_a$ denote the video and audio token rates, respectively.
This rate-aware rescaling places audio and video tokens into a more comparable temporal coordinate system for joint attention.

Context is injected separately through \emph{Context-Guided Cross-Attention \& FFNs}.
This preserves a dedicated audio-video synchronization space while allowing textual and timbre conditions to modulate the denoising trajectory.
Compared with fully unified tri-modal attention, this design avoids inserting context tokens directly into the same self-attention space used for low-level audio-video synchronization.

\paragraph{Unified Fusion Layers.}
After audio-video correspondence has been established, NAVA transitions to \emph{Unified Fusion Layers}.
In these layers, audio and video tokens are processed with \emph{Modality-Shared Unified Projection} and updated by shared transformer blocks.
Since the preceding alignment layers have already reduced the representational gap between audio and video tokens, parameter sharing in later layers becomes more stable and efficient.
This removes persistent stream separation and encourages compact collaborative denoising in a shared generation space.
Context remains external through cross-attention, so semantic guidance and controllable conditions continue to modulate the joint denoising process without disrupting the learned synchronization structure.

\subsection{Timbre-in-Context Conditioning}

Textual context provides semantic guidance, while speech-driven audio-video generation further requires segment-level timbre control, i.e., specifying \emph{who speaks which content}.
We propose \emph{Timbre-in-Context Conditioning}, which represents reference timbre cues as context tokens and binds them to their corresponding speech spans through the existing context-conditioning pathway.

Let $\mathcal{P}$ denote the textual prompt containing speech spans $\{\mathcal{S}_i\}_{i=1}^{N}$, and let $\mathcal{R}_i$ be the reference utterance specifying the desired timbre for $\mathcal{S}_i$.
We extract a context-space timbre token as
\begin{equation}
    \mathbf{s}_i = E_{\mathrm{tim}}(\mathcal{R}_i),
\end{equation}
where $E_{\mathrm{tim}}$ denotes the timbre encoder.
Each speech span is then augmented as
\begin{equation}
    \mathcal{S}_i
    \rightarrow
    \big[
    \langle \mathrm{S} \rangle,\,
    \mathbf{s}_i,\,
    \mathrm{Text}(\mathcal{S}_i),\,
    \langle \mathrm{E} \rangle
    \big],
\end{equation}
where $\langle \mathrm{S} \rangle$ and $\langle \mathrm{E} \rangle$ mark the boundaries of a timbre-conditioned speech span.
Applying this replacement to all speech spans yields the final context sequence:
\begin{equation}
    \mathbf{c}
    =
    \mathrm{Augment}
    \left(
    \mathcal{P};
    \{(\mathcal{S}_i, \mathbf{s}_i)\}_{i=1}^{N}
    \right).
\end{equation}

During denoising, NAVA accesses this augmented context through context-guided cross-attention.
Thus, timbre cues are associated with speech spans within the original prompt structure rather than injected as a global control signal.
This is important for multi-speaker generation, where different utterances may require different speaker identities or timbre styles.
Because timbre information is represented in the context pathway, the mechanism requires no auxiliary speaker-control branch or backbone modification.
It naturally supports compositional control by assigning different timbre tokens to different speech spans, while keeping the audio-video denoising backbone unchanged.

\subsection{Training and Inference}
\label{sec:training_inference}

\subsubsection{Progressive Multi-Task Training}

NAVA is trained with a progressive multi-task strategy over T2AV, TI2AV, T2A, T2V, and TIA2AV tasks, covering audio-only, video-only, and paired audio-visual denoising trajectories.
The training schedule consists of three stages.
First, we train on audio-only and paired audio-visual data with a $3{:}1$ sampling ratio to initialize the audio pathway and stabilize audio denoising while preserving the visual capability inherited from the pretrained video backbone~\cite{wan2025wan}.
We then shift the audio-only/audio-visual ratio to $1{:}2$ and train on high-quality audio data together with the full audio-visual dataset to improve audio fidelity and audio-visual synchronization.
Finally, we fine-tune on curated high-quality audio-visual data to improve instruction following and controllable generation, including multi-speaker dialogue, complex motion, and camera control.

\subsubsection{Structured Dropout for Guidance}

To support condition-factorized guidance, we construct paired conditional and partially unconditional denoising paths during training, enabling guidance signals to be estimated from controlled prediction differences.
For audio-visual alignment, we apply \emph{Random Cross-modality Attention Masking}, where cross-modal attention entries between audio and video tokens are randomly masked while intra-modal attention remains intact.
This exposes the model to both coupled and partially decoupled audio-video denoising regimes, whose prediction contrast is later used for alignment guidance.
For timbre control, we apply \emph{Random Timbre-in-Context Conditioning} by dropping or replacing timbre tokens with null tokens for a subset of speech spans.
This trains the model under timbre-conditioned and timbre-free contexts, providing the prediction contrast required for timbre guidance.

\subsubsection{Condition-Factorized Classifier-Free Guidance}

During inference, we build on the audio-visual guidance formulation of LTX~\cite{hacohen2026ltx} and extend it with reference-timbre guidance.
Let $\mathbf{v}_\theta^{c,a,\tau}(z_t)$ denote the prediction at step $t$, where $z_t$ is the noisy audio-video latent, and $c$, $a$, and $\tau$ denote textual context, audio-video interaction, and reference timbre conditioning, respectively.
We define three guidance directions:
\begin{equation}
\begin{aligned}
    \Delta_{\mathrm{text}}
    &=
    \mathbf{v}_\theta^{c,a,\tau}(z_t)
    -
    \mathbf{v}_\theta^{\varnothing,a,\tau}(z_t), \\
    \Delta_{\mathrm{align}}
    &=
    \mathbf{v}_\theta^{c,a,\tau}(z_t)
    -
    \mathbf{v}_\theta^{c,\varnothing,\tau}(z_t), \\
    \Delta_{\mathrm{timbre}}
    &=
    \mathbf{v}_\theta^{c,a,\tau}(z_t)
    -
    \mathbf{v}_\theta^{c,a,\varnothing}(z_t).
\end{aligned}
\end{equation}
The final guided prediction is
\begin{equation}
    \hat{\mathbf{v}}_\theta(z_t)
    =
    \mathbf{v}_\theta^{c,a,\tau}(z_t)
    +
    s_{\mathrm{text}}\Delta_{\mathrm{text}}
    +
    s_{\mathrm{align}}\Delta_{\mathrm{align}}
    +
    s_{\mathrm{timbre}}\Delta_{\mathrm{timbre}},
\end{equation}
where $s_{\mathrm{text}}$, $s_{\mathrm{align}}$, and $s_{\mathrm{timbre}}$ control prompt adherence, audio-visual synchronization, and timbre preservation, respectively.
This factorized formulation supports decoupled alignment guidance and fine-grained timbre control during inference.
\section{Experiments}

\subsection{Experimental Setup}

\paragraph{Implementation details.}
NAVA has 6.3B parameters with 30 MMDiT blocks, where the first 10 blocks are \emph{Hierarchical Alignment Layers} and the remaining 20 are \emph{Unified Fusion Layers}.
We initialize corresponding layers from Wan2.2-5B~\cite{wan2025wan}, use Wan2.2-VAE for video latents with a $4\times16\times16$ compression ratio, and use LTX2.3-VAE for multi-channel audio latents.
The model is trained with AdamW at a learning rate of $5\times10^{-5}$ on 128 NVIDIA H100 GPUs, with an effective batch size of 512 for 70K steps following the three-stage schedule in Sec.~\ref{sec:training_inference}.
We apply random cross-modality attention masking and timbre-condition dropout with probabilities of $20\%$ each, and sample image conditions with probability $50\%$.

\paragraph{Benchmarks and baselines.}
Following MoVA~\cite{team2026mova} and daVinci-MagiHuman~\cite{chern2026speed}, we adopt Verse-Bench~\cite{wang2025universe} for objective audio-visual evaluation, covering speech videos, sound effects, and musical instruments.
We further evaluate timbre controllability on the Seed-TTS benchmark~\cite{anastassiou2024seed}.
For Verse-Bench, we compare with Ovi-1.1~\cite{low2025ovitwinbackbonecrossmodal}, MoVA~\cite{team2026mova}, LTX-2.3~\cite{hacohen2026ltx}, and daVinci-MagiHuman~\cite{chern2026speed}, covering dual-tower and tri-modal unified paradigms.
For Seed-TTS, we compare with DreamID-Omni~\cite{guo2026dreamid}.
Since DreamID-Omni requires paired reference audio and image inputs, we use a fixed reference image for all samples and provide the corresponding reference audio.
For fair comparison, we evaluate the base version of each model without additional super-resolution, distillation, or post-processing modules.
We also apply Gemini-3-Flash rewriting to all test prompts to match each model's expected inference format while preserving the original benchmark semantics.

\paragraph{Evaluation metrics.}
We evaluate the proposed method along four dimensions: audio--visual alignment, video quality, audio quality, and timbre controllability, covering both perceptual fidelity and cross-modal consistency.
For audio--visual alignment, we report Sync-C and Sync-D from SyncNet~\cite{chung2016out}, which measure the confidence and temporal offset of lip--audio synchronization, respectively. We further use the ImageBind score (IB-Score)~\cite{girdhar2023imagebind} to assess cross-modal semantic consistency between the generated video and audio.
For video quality, we report identity consistency and aesthetic score.
For audio quality, we employ Audiobox-Aesthetics~\cite{tjandra2025aes}, a no-reference audio assessment model trained to predict human perceptual judgments along multiple aesthetic axes. Specifically, we report Production Quality (PQ) to assess perceived audio fidelity, and Fréchet Distance (FD) to measure the distributional gap between generated and reference audio in the learned audio feature space. In addition, we report word error rate (WER) to measure speech intelligibility and content accuracy.
For timbre controllability, we compute Seed-TTS timbre similarity between the generated speech and the reference utterance.
Higher values are better for Sync-C, IB-Score, video quality, PQ, and timbre similarity, whereas lower values are better for Sync-D, FD, and WER.

\subsection{Main Results}
\begin{table}[t]
\centering
\caption{
\textbf{General capabilities on Verse-Bench.}
We compare representative audio-video generation models in terms of synchronization, video quality, and audio quality.
NAVA achieves the strongest overall synchronization and video quality, with competitive audio quality and the fewest parameters.
}
\label{tab:main}
\resizebox{\textwidth}{!}{
\begin{tabular}{lccccccccc}
\toprule
\multirow{2}{*}{\textbf{Model}} 
& \multirow{2}{*}{\textbf{Params}} 
& \multirow{2}{*}{\textbf{Resolution}}
& \multicolumn{3}{c}{\textbf{AV-Align}}
& \multirow{2}{*}{\textbf{Video Quality $\uparrow$}}
& \multicolumn{3}{c}{\textbf{Audio}} \\
\cmidrule(lr){4-6} \cmidrule(lr){8-10}
& & 
& \textbf{Sync-C $\uparrow$} 
& \textbf{Sync-D $\downarrow$} 
& \textbf{IB $\uparrow$}
& 
& \textbf{WER $\downarrow$} 
& \textbf{PQ $\uparrow$} 
& \textbf{FD $\downarrow$} \\
\midrule

Ovi 1.1~\cite{low2025ovitwinbackbonecrossmodal}
& 10B
& 720p
& \underline{7.484}
& 7.979
& 0.199
& \underline{0.636}
& 0.102
& 5.843
& 0.942 \\

MOVA~\cite{team2026mova}
& A18B (32B)
& 720p
& 7.289
& 7.808
& 0.269
& 0.603
& 0.126
& \textbf{7.233}
& 0.922 \\

Davinci~\cite{chern2026speed}
& 15B
& 540p
& 7.149
& 7.816
& 0.269
& 0.600
& 0.151
& 5.956
& 0.931 \\

LTX 2.3~\cite{hacohen2026ltx}
& 19B
& 512p
& 7.248
& \underline{7.690}
& \textbf{0.337}
& 0.576
& 0.106
& \underline{6.946}
& \textbf{0.829} \\

\textbf{NAVA} (ours)
& \textbf{6.3B}
& \textbf{720p}
& \textbf{7.791}
& \textbf{7.566}
& \underline{0.313}
& \textbf{0.659}
& \textbf{0.099}
& 6.861
& \underline{0.833} \\

\bottomrule
\end{tabular}
}
\vspace{0.5em}
\end{table}
\begin{table}[t]
\centering
\caption{
    \textbf{Reference-timbre generation on Seed-TTS.}
    We report WER and speaker similarity for both audio-only speech models and audio-video generation models under an audio-only evaluation protocol.
    NAVA achieves the best results among audio-video generation models.
}
\label{tab:sim}
\resizebox{0.62\textwidth}{!}{
\begin{tabular}{llcc}
\toprule
\textbf{Model Category} & \textbf{Model} & \textbf{WER $\downarrow$} & \textbf{Speaker Similarity $\uparrow$} \\
\midrule

\multirow{3}{*}{Audio}
& CosyVoice~\cite{du2024cosyvoice}      & 4.29  & 60.9 \\
& CosyVoice2~\cite{du2024cosyvoice2}   & 2.57  & 65.2 \\
& Qwen2.5-Omni~\cite{Qwen2.5-Omni}   & 2.72  & 63.2 \\
\midrule

\multirow{2}{*}{Audio-Video}
& DreamID-Omni~\cite{guo2026dreamid}   & 31.76 & 35.7 \\
& NAVA           & \textbf{4.20} & \textbf{66.7} \\

\bottomrule
\end{tabular}
}
\end{table}
\begin{table}[t]
\centering
\caption{
\textbf{Ablation of Align-then-Fuse MMDiT.}
We compare model variants with different combinations of \emph{Hierarchical Alignment Layers} (HAL) and \emph{Unified Fusion Layers} (UFL).
Results demonstrate that combining HAL and UFL yields the best alignment and video quality, while maintaining competitive audio quality.
}
\label{tab:ablation_layers}
\resizebox{0.72\textwidth}{!}{
\begin{tabular}{cccccccc}
\toprule
\multirow{2}{*}[-0.5ex]{\textbf{\makecell{HAL\\Layers}}}
& \multirow{2}{*}[-0.5ex]{\textbf{\makecell{UFL\\Layers}}}
& \multirow{2}{*}[-0.5ex]{\textbf{\makecell{Model\\Params}}}
& \multicolumn{2}{c}{\textbf{Alignment}}
& \multirow{2}{*}[-0.5ex]{\textbf{\makecell{Video\\Quality $\uparrow$}}}
& \multicolumn{2}{c}{\textbf{Audio Quality}} \\
\cmidrule(lr){4-5} \cmidrule(lr){7-8}
& &
& \textbf{Sync-C $\uparrow$}
& \textbf{IB $\uparrow$}
&
& \textbf{PQ $\uparrow$}
& \textbf{WER $\downarrow$} \\
\midrule

& \(\circ\)
& 5B
& 7.643
& 33.22
& 67.53
& 5.296
& 0.182 \\

\(\circ\)
& \(\circ\)
& 6.3B
& \textbf{7.684}
& \textbf{34.34}
& \textbf{67.67}
& \textbf{5.377}
& 0.177 \\

\(\circ\)
&
& 7.7B
& 7.030
& 30.91
& 66.62
& 5.347
& \textbf{0.167} \\

\bottomrule
\end{tabular}
}
\end{table}

\subsubsection{Quantitative Evaluation}

Table~\ref{tab:main} reports quantitative results on Verse-Bench.
NAVA achieves the best overall trade-off across audio--visual alignment, video quality, audio quality, and model efficiency.
With only $6.3$B parameters, NAVA obtains the highest Sync-C score of $7.791$ and the lowest Sync-D score of $7.566$, demonstrating superior temporal synchronization between generated speech and visual motion.
It also achieves the best video quality score of $0.659$, suggesting that the proposed Align-then-Fuse design preserves strong visual generation capability while enabling synchronized audio generation.

For semantic audio--visual consistency, NAVA obtains an IB-Score of $0.313$, outperforming Ovi-1.1 and remaining competitive with MoVA and Davinci, although LTX~2.3 achieves the highest IB-Score.
For audio quality, NAVA achieves the lowest WER of $0.099$, indicating improved speech intelligibility and content accuracy.
Its PQ and FD scores, $6.861$ and $0.833$, are also competitive among baselines, showing that NAVA maintains high perceived audio fidelity and a close distributional match to reference audio.
These results indicate that NAVA substantially improves audio--visual synchronization and video quality without sacrificing audio quality, despite using the fewest parameters among the compared audio-video models.

Table~\ref{tab:sim} evaluates reference-timbre speech generation on the EN subset of the Seed-TTS benchmark.
Audio-only speech models such as CosyVoice~\cite{du2024cosyvoice}, CosyVoice2~\cite{du2024cosyvoice2}, and Qwen2.5-Omni~\cite{Qwen2.5-Omni} provide strong references for pure speech generation.
Despite operating as an audio-video generation model with synchronized visual generation, NAVA achieves the highest speaker similarity of $66.7$ and a competitive WER of $4.20$.
Within the audio-video model category, NAVA substantially outperforms DreamID-Omni, reducing WER from $31.76$ to $4.20$ and improving speaker similarity from $35.7$ to $66.7$.
These results demonstrate the effectiveness of \emph{Timbre-in-Context Conditioning}, which binds reference timbre cues to corresponding speech spans through the context pathway.
Overall, NAVA provides a strong balance across synchronization, semantic consistency, video quality, audio quality, and timbre controllability.

\subsubsection{Qualitative Evaluation}

\paragraph{Visualization.}
\begin{figure*}[t]
    \centering
    \includegraphics[width=\textwidth]{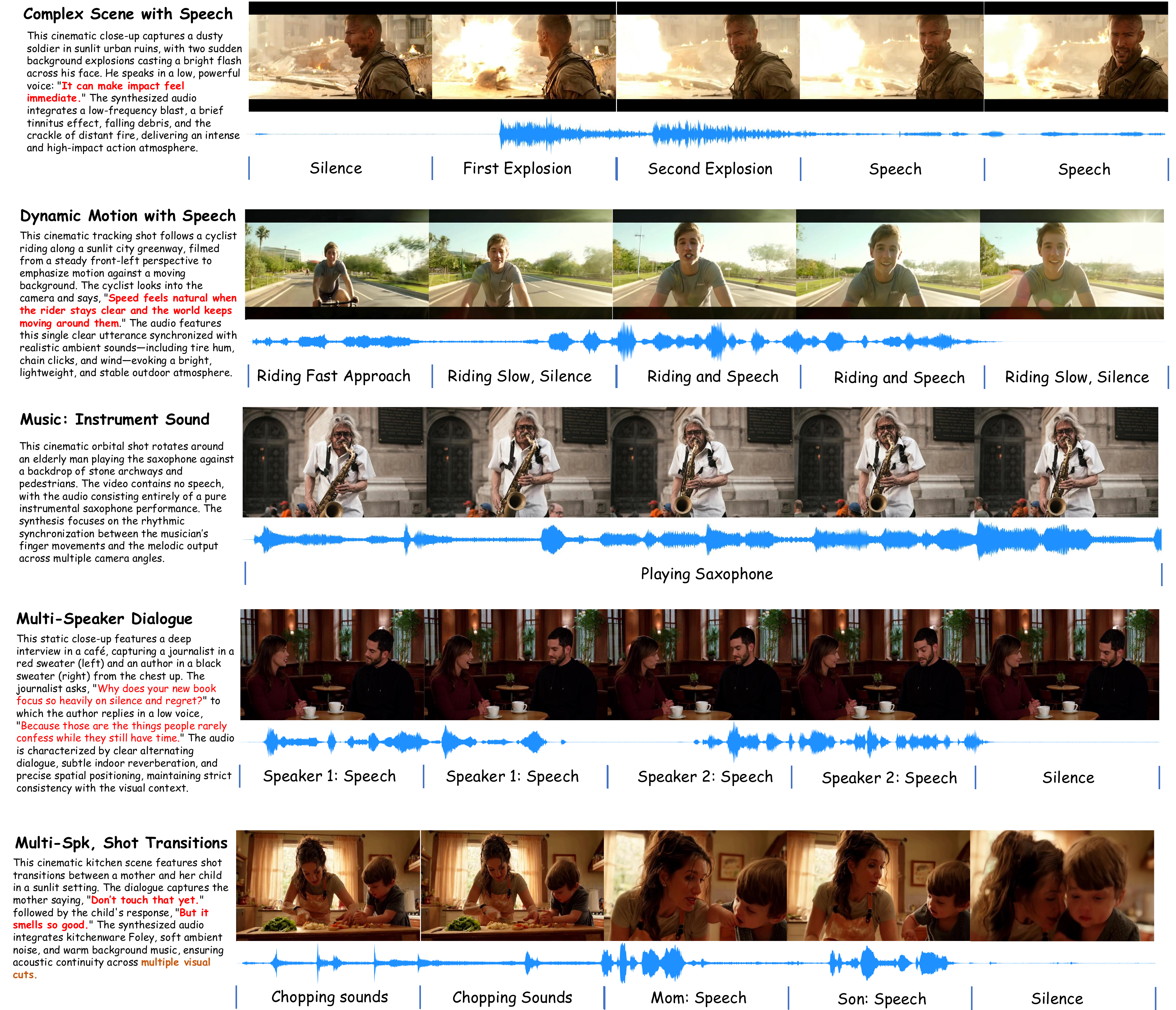}
    \caption{
\textbf{Qualitative visualization of NAVA.}
We present various generated video frames, audio waveforms, and event-level annotations across diverse scenarios, including complex speech scenes, dynamic motion, musical performance, multi-speaker dialogue, and shot transitions.
The annotations highlight how the generated sounds are temporally aligned with visual events, such as silence, explosions, riding motion, instrumental performances, speaker turns, chopping sounds, and scene cuts.
Overall, NAVA produces semantically coherent audio-visual outputs with diverse sound events, scene-aware acoustics, and controllable speaker assignment.
}
    \label{fig:qua}
\end{figure*}
Fig.~\ref{fig:qua} visualizes representative NAVA generations across challenging scenarios, including speech in complex acoustic scenes, speech during dynamic motion, musical performance, multi-speaker dialogue, and shot transitions.
The sampled frames, waveforms, and event annotations show that NAVA can synthesize temporally synchronized speech, sound effects, and instrumental audio under complex visual contexts.
The examples also demonstrate controllable speaker assignment and coherent generation across multi-speaker and multi-shot settings.

\paragraph{User Study.}

To further assess perceptual quality and robustness, we conduct a human evaluation using the GSB protocol.
We evaluate 250 cases covering both text-to-audio-video (T2AV) and text-image-to-audio-video (TI2AV) generation.
For T2AV, we construct a diverse set of synthetic prompts to cover challenging scenarios, including single- and dual-speaker speech, camera control, ambient sound, musical instruments, and complex acoustic events.For TI2AV, we directly use samples from Verse-Bench. MoVA is excluded from the T2AV comparison because its released model is not designed to take text-only inputs for this generation mode.
To obtain a more fine-grained and reliable evaluation, participants are asked to compare paired results along two dimensions: overall audio-visual quality and audio-visual alignment accuracy.
For each pair, participants assign one of three preferences: Win, Tie, or Lose.

\begin{figure*}[t]
    \centering
    \includegraphics[width=\textwidth]{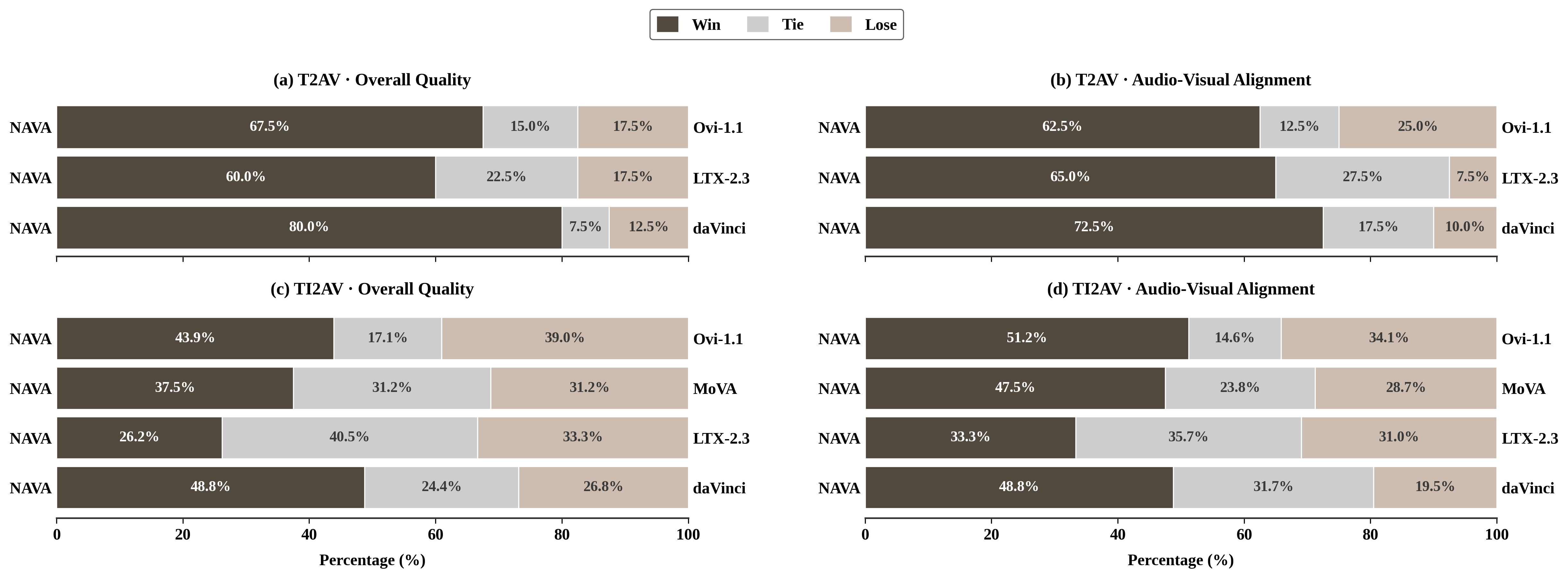}
    \caption{
    \textbf{Results of User study.}
    Pairwise human preference comparisons between NAVA and representative baselines under T2AV and TI2AV settings.
    The bars report the win/tie/lose percentages of NAVA in terms of overall quality and audio-visual alignment.
    NAVA achieves favorable preferences in most comparisons, especially on audio-visual alignment.
    }
    \label{fig:user_study}
\end{figure*}

The results are shown in Fig.~\ref{fig:user_study}.
On T2AV, NAVA consistently outperforms all compared baselines.
In terms of overall quality, NAVA achieves win rates of $67.5\%$, $60.0\%$, and $80.0\%$ against Ovi-1.1, LTX-2.3, and daVinci, respectively.
For audio-visual alignment, NAVA obtains even stronger preferences, with win rates of $62.5\%$, $65.0\%$, and $72.5\%$.
These results suggest that NAVA generalizes well to diverse and challenging text-driven audio-visual generation scenarios, especially in maintaining synchronized speech, motion, and sound events.

On TI2AV, NAVA also shows clear advantages over most baselines.
For overall quality, NAVA achieves win rates of $43.9\%$, $37.5\%$, $26.2\%$, and $48.8\%$ against Ovi-1.1, MoVA, LTX-2.3, and daVinci, respectively.
For audio-visual alignment, NAVA obtains win rates of $51.2\%$, $47.5\%$, $33.3\%$, and $48.8\%$.
While LTX-2.3 remains competitive on TI2AV, particularly in overall quality, NAVA achieves stronger alignment against Ovi-1.1, MoVA, and daVinci, and remains competitive with LTX-2.3.
Overall, the human evaluation confirms that NAVA provides favorable perceptual audio-visual quality and more reliable temporal alignment across both T2AV and TI2AV settings.

\subsection{Ablation Studies}

\begin{table}[t]
\centering
\caption{
\textbf{Ablation of condition-factorized CFG.}
Alignment CFG is evaluated on Verse-Bench, while timbre CFG is evaluated on the Seed-TTS benchmark.
% Best results within each group are marked in \textbf{bold}.
}
\label{tab:cfg_ablation}
\setlength{\tabcolsep}{4.5pt}
\resizebox{\textwidth}{!}{
\begin{tabular}{lcccccccc}
\toprule
% \textbf{Variant}
& \textbf{Sync-C $\uparrow$}
& \textbf{Sync-D $\downarrow$}
& \textbf{IB $\uparrow$}
& \textbf{Video Quality $\uparrow$}
& \textbf{PQ $\uparrow$}
& \textbf{WER$_{\mathrm{VB}}$ $\downarrow$}
& \textbf{WER$_{\mathrm{Seed}}$ $\downarrow$}
& \textbf{ASV $\uparrow$} \\
\midrule
\multicolumn{9}{c}{\textbf{Alignment CFG on Verse-Bench}} \\
\midrule
NAVA w/o Align CFG
& 6.170
& 8.755
& 0.355
& \textbf{0.667}
& 6.658
& 0.126
& --
& -- \\

NAVA w/ Align CFG
& \textbf{7.791}
& \textbf{7.566}
& \textbf{0.402}
& 0.659
& \textbf{6.860}
& \textbf{0.099}
& --
& -- \\

\midrule
\multicolumn{9}{c}{\textbf{Timbre CFG on Seed-TTS}} \\
\midrule
NAVA w/o Timbre CFG
& --
& --
& --
& --
& --
& --
& \textbf{3.78}
& 65.5 \\

NAVA w/ Timbre CFG
& --
& --
& --
& --
& --
& --
& 4.20
& \textbf{66.7} \\

\bottomrule
\end{tabular}
}
\end{table}

\paragraph{Ablation on Align-then-Fuse MMDiT.}
Table~\ref{tab:ablation_layers} studies the roles of \emph{Hierarchical Alignment Layers}~(HAL) and \emph{Unified Fusion Layers}~(UFL).
The UFL-only variant removes early modality-aware alignment and directly shares generation parameters across audio and video, leading to weaker Sync-C and IB scores.
This suggests that fully shared generation without an explicit alignment stage is insufficient for establishing fine-grained audio-video correspondence.
The HAL-only variant achieves stronger audio-related metrics, including improved speech intelligibility and audio fidelity, but degrades IB and video quality, indicating that persistent modality-aware alignment can preserve unimodal denoising ability at the cost of compact high-level fusion.
Combining HAL and UFL achieves the best overall trade-off, obtaining the strongest synchronization, cross-modal consistency, and video quality.
These results support the proposed align-then-fuse design: HAL first aligns heterogeneous audio-video representations, while UFL promotes shared high-level generation and collaborative denoising after correspondence has been established.

\paragraph{Ablation on Condition-Factorized CFG.}
Table~\ref{tab:cfg_ablation} evaluates the proposed condition-factorized classifier-free guidance.
Alignment CFG substantially improves audio-visual correspondence, increasing Sync-C from $6.170$ to $7.791$, reducing Sync-D from $8.755$ to $7.566$, and improving IB from $0.355$ to $0.402$.
These gains come with only minor changes in video quality, while WER decreases from $0.126$ to $0.099$ and PQ improves from $6.658$ to $6.860$, showing that alignment guidance strengthens synchronization without compromising unimodal generation quality.
For timbre control, Timbre CFG improves reference-timbre consistency, increasing ASV from $65.5$ to $66.7$, with a mild WER trade-off from $3.78$ to $4.20$.
Overall, these ablations show that different guidance directions modulate distinct generation attributes: Alignment CFG improves audio-video correspondence, while Timbre CFG improves timbre consistency.
This validates factorized guidance for separately adjusting synchronization and timbre controllability at inference time, without retraining or modifying the generation backbone.

\section{Related Work}

\paragraph{Video-to-Audio Generation.}
Video-to-audio generation synthesizes acoustic content conditioned on a given video, and often serves as a cascaded component for audio-visual content creation.
Early methods explore multimodal representation learning and cross-modal conditioning, using Transformer architectures or visual-textual encoders to fuse video and text cues~\cite{akbari2021vatt, hu2024video}.
Recent systems improve temporal precision and generation efficiency through high-frame-rate visual features, rectified flow matching, and large-scale audio-visual training~\cite{viertola2025temporally, wang2024frieren}.
More recent works such as MMAudio~\cite{cheng2025mmaudio} and Kling-Foley~\cite{wang2025kling} adopt diffusion or MMDiT-style architectures and leverage large-scale video-audio corpora such as VGGSound~\cite{chen2020vggsound} and WavCaps~\cite{mei2024wavcaps}.
Although these approaches can generate plausible audio for existing videos, they are inherently conditioned on a fixed visual trajectory and do not address native joint generation, where audio and video should co-evolve during synthesis.

\paragraph{Audio-Video Joint Generation.}
Unlike video-to-audio generation, audio-video joint generation synthesizes both modalities within a shared generation process and therefore requires tighter temporal and semantic coordination.
Early attempts such as MM-Diffusion~\cite{ruan2023mm}, Javis-DiT~\cite{liu2025javisdit}, and Universe-1~\cite{wang2025universe} explore cross-modal attention, expert composition, or multimodal diffusion for coordinated generation.
Recent open-source systems, including UniAVGen~\cite{zhang2025uniavgen}, Ovi~\cite{low2025ovitwinbackbonecrossmodal}, LTX~\cite{hacohen2026ltx}, and MoVA~\cite{team2026mova}, mainly adopt dual-tower designs that maintain separate audio and video streams and introduce posterior fusion or alignment.
Such designs can exploit pretrained unimodal priors, but delayed audio-video interaction may limit fine-grained synchronization and semantic consistency.

Unified modeling has been explored to strengthen cross-modal interaction.
Apollo~\cite{wang2026klear} applies joint attention over concatenated multimodal tokens, while daVinci-MagiHuman~\cite{chern2026speed} places textual context, video, and audio tokens into a shared tri-modal space.
These designs enable direct multimodal interaction, but fully mixing semantic context with generation modalities can entangle high-level conditioning with low-level audio-video synchronization.
In contrast, NAVA establishes audio-video correspondence in a dedicated interaction space and injects context as external conditioning, separating native synchronization from semantic and controllable guidance.

\paragraph{Controllable Audio-Visual Generation.}
Controllable audio-visual generation requires not only synchronized audio and video, but also flexible conditioning on identity, reference audio, speaker style, or timbre.
UniAVGen~\cite{zhang2025uniavgen} and DreamID-Omni~\cite{guo2026dreamid} incorporate reference tokens or identity/timbre conditions to support controllable generation.
However, many reference-conditioning mechanisms are applied as global controls or auxiliary branches, which can be less flexible for multi-speaker scenarios where different utterances require different timbres.
NAVA instead represents reference timbre cues as context tokens tied to specific speech spans.
This enables compositional content-timbre binding through the existing context-conditioning pathway, without introducing an additional speaker-control branch or modifying the denoising backbone.

\section{Conclusion}

We presented \textbf{NAVA}, a Native Audio-Visual Alignment framework for joint audio-video generation.
NAVA decouples audio-visual synchronization from context conditioning by establishing audio-video correspondence in a dedicated alignment space and using context as external guidance.
We instantiate this formulation with an \emph{Align-then-Fuse MMDiT} architecture, which bridges modality-aware alignment and unified audio-video denoising, and introduce \emph{Timbre-in-Context Conditioning} for segment-level reference-timbre control.
Experiments on Verse-Bench and Seed-TTS demonstrate that NAVA achieves strong audio-visual synchronization, visual quality, semantic consistency, and timbre controllability.
These results indicate that native audio-visual alignment with decoupled context conditioning is a promising direction for scalable and controllable audio-video generation.

% \paragraph{Limitations and Future Work.}
Despite its strong overall performance, NAVA remains limited in generating certain long-tail and highly compositional audio events, such as rare animal sounds, music, singing, and complex mixtures of scene sounds.
Addressing these limitations requires broader and more meticulously curated audio-visual data, especially for rare events and compositionally rich scenarios.
Our results suggest that deeper audio-visual coupling represents a highly promising direction.
In the future, another potential direction is to explore earlier fusion mechanisms, such as joint audio-visual tokenizers or unified representation models, to further enhance synchronization, semantic consistency, and generalization.

\bibliographystyle{unsrtnat}
\bibliography{references}

\newpage
\section{Appendix}

\subsection{Data Pipeline}
\textbf{Large-scale collection and preprocessing.}
We construct a large-scale audio-visual training corpus from heterogeneous sources, including Koala-36M, TED-style speech videos, and raw movie/TV footage. The raw videos are first segmented at scale with a Hadoop-based pipeline. To improve data quality and reduce shortcut learning from overlaid text, we apply OCR-based filtering and subtitle removal using PaddleOCR. We further remove redundant or near-duplicate clips by extracting video embeddings with VideoCLIP and performing large-scale k-means clustering, followed by category-level merging and filtering of small clusters.

\textbf{Modality-aware tagging and subset construction.}
We then annotate each clip with both visual and acoustic metadata. For visual content, we use VLM-based filtering and tagging to retain clips with clear visual quality and coherent events or transitions, and assign semantic tags such as movies, documentaries, TV series, live streams, speeches, news, and interviews. For audio content, we combine YAMNet-based audio classification with an omni-modal tagger to categorize clips into single-speaker speech, multi-speaker speech, ambient sound, music, and singing. These tags are used to construct different data subsets for pretraining and supervised fine-tuning.

\textbf{Hierarchical audio-visual caption annotation.}
For caption annotation, we adopt a two-stage strategy. On the full-scale dataset, video and audio captions are generated separately using Qwen3-VL and Qwen3-Omni, and are then fused by either direct concatenation or rewriting by Gemini-3-Flash. For high-quality and multi-speaker subsets, we use Gemini-3-Pro to produce more accurate, structured, and temporally grounded audio-visual captions.

\textbf{Multi-operator quality filtering.}
Finally, we apply a set of quality assessment operators to filter low-quality samples. These operators cover visual quality, including aesthetics, sharpness, brightness, and motion score; audio quality, including AudioBox Aesthetic scores; and audio-visual alignment, measured by SyncNet, SyncFormer, and ImageBind. The resulting corpus consists of diverse, deduplicated, high-quality, and richly annotated audio-visual clips, supporting scalable multi-stage audio-visual model training.

\subsection{Data Statistics}
Our raw collection contains approximately 20M audio clips and 100M video clips. After subtitle filtering, quality filtering, near-duplicate removal, and audio-visual alignment filtering, we obtain around 15M clips for large-scale training. Koala-36M contributes approximately 20\% of the final training corpus. The average video duration is about 7 seconds. For the supervised fine-tuning stage, we further apply multi-operator collaborative filtering and retain 160K high-quality samples with accurate captions and strong audio-visual alignment.

\subsection{Prompt Engineering}

Audio-visual generation is conditioned on multiple intertwined factors,
including scene layout, subject appearance, motion, camera behavior, lighting,
style, speech, speaker characteristics, environmental sounds, music, and
spatial acoustics. Unlike purely visual generation, audio-visual generation
requires the prompt to specify not only what appears in the video, but also what
is heard and how the audible events are temporally aligned with the visual
dynamics. Therefore, we use structured dense captions rather than free-form
short descriptions for both training and inference.

We design a unified prompt rewriting template that decomposes each video into
global visual semantics, temporal dynamics, camera and composition, and audio
events. The audio branch is compatible with non-speech scene sounds,
single-speaker speech, multi-speaker dialogue, music, singing, and ambient
audio. For speech videos, utterances are explicitly marked with
\texttt{<S>} and \texttt{<E>} tokens, while speaker timbre, emotion, speaking
rate, and sound-field position are also described. For non-speech videos, the
template emphasizes action sounds, contact and friction sounds, object sounds,
environmental ambience, and reverberation. This design yields captions that are
visually detailed, temporally ordered, acoustically grounded, and consistent
across heterogeneous audio-visual data.

\paragraph{Audio-Visual Prompt Template.}The full template used for prompt rewriting is shown below.
\begin{tcolorbox}[
    title=Unified Audio-Visual Caption Prompt Template,
    colback=gray!3,
    colframe=gray!45,
    boxrule=0.5pt,
    arc=2pt,
    left=4pt,
    right=4pt,
    top=4pt,
    bottom=4pt
]
\small
\begin{verbatim}
Describe the video in a structured audio-visual caption. The caption should
cover visual content, temporal dynamics, camera behavior, and all audible
events. Use natural language, but preserve the temporal order of the video.

[Global visual description]
This is a [style/genre] video with a [overall mood/atmosphere] feeling.
The main subject(s) are [subject(s)] located in [scene/environment].
Describe the subject appearance, posture, facial expression, clothing,
materials, objects, and spatial arrangement. Also describe the background
elements, lighting condition, dominant color tone, and environmental mood.

[Temporal dynamics]
At the beginning of the video, [initial visual state]. Then, [action or event
development 1]. Next, [action or event development 2]. As the video continues,
describe [motion details], [object interactions], and [environmental feedback].
At the end of the video, [ending state], producing [final visual/audio-visual
effect].

[Camera and composition]
Describe the camera angle, shot scale, framing, and composition. Specify
whether the camera is fixed, handheld, tracking, panning, tilting, zooming, or
pushing in. Mention whether there are cuts or scene transitions. The caption
should highlight how the camera captures [key action], [subject details], and
[environmental dynamics].

[Audio description]
Describe all audible content in temporal order. The audio may contain
non-speech sounds, single-speaker speech, multi-speaker dialogue, music,
singing, or ambient sound. Always describe the relationship between sound and
visual events when possible.

If there is no speech:
    State that there is no spoken dialogue or voice-over.
    Describe the main sound effects, including [action sounds],
    [contact/friction sounds], [object sounds], [environmental sounds],
    [music or singing if present], [spatial echo/reverberation], and
    [overall auditory impression].

If there is single-speaker speech:
    Describe the speaker's action, expression, position, voice timbre,
    emotion, speaking rate, and sound-field position.
    Transcribe the utterance using:
    <S>[spoken content]<E>
    Also describe the background ambience, object sounds, reverberation,
    and whether music is present.

If there is multi-speaker dialogue:
    For each speaker turn, describe the speaker's action, expression,
    voice timbre, emotion, speaking rate, and sound-field position.
    Transcribe each utterance using:
    [Speaker A description]: <S>[utterance A]<E>
    [Speaker B description]: <S>[utterance B]<E>
    Continue this format for additional speakers or turns.
    Also describe turn-taking, speaker interaction, background ambience,
    object sounds, reverberation, and mixing quality.
\end{verbatim}

\end{tcolorbox}

\paragraph{Example Captions.}
Representative rewritten captions are shown below. These examples demonstrate
that the proposed template can consistently capture visual details, temporal
evolution, camera behavior, dialogue structure, speaker-specific acoustic
properties, and background sound events.

\begin{tcolorbox}[
    title=Example: Compact Dialogue Caption,
    colback=gray!3,
    colframe=gray!45,
    boxrule=0.5pt,
    arc=2pt,
    left=4pt,
    right=4pt,
    top=4pt,
    bottom=4pt
]
\small
This realistic animated video shows a two-person conversation in a warm coffee
shop. The young man on the left wears a dark blue hoodie and looks curious,
while the older man on the right wears round glasses, has a white beard, holds
a coffee cup with both hands, and appears calm and gentle. They sit face-to-face
at a wooden table, both framed from the chest up. The background contains a
coffee menu board, warm pendant lights, plants, and soft window light.

At the beginning of the video, the young man on the left asks clearly:
\texttt{<S>Can I use any timbre I want?<E>}
The older man on the right nods gently and responds in a low and warm voice:
\texttt{<S>Yes, in all kinds of complex scenes.<E>}
The video then ends.

The camera remains fixed in a medium-close shot, with no cuts, zooms, pans, or
tilts. The audio contains soft cafe ambience, subtle cup contact sounds, and
indoor reverberation. The two voices alternate clearly with accurate left-right
sound-field positioning, highlighting speaker identity, timbre contrast, and a
natural conversational rhythm.
\end{tcolorbox}

\paragraph{Infrastructure and Training Cost.}
We train NAVA with a distributed infrastructure designed for large-scale, long-context audio-visual modeling.
To reduce memory overhead, we adopt Fully Sharded Data Parallel (FSDP), which shards model parameters, gradients, and optimizer states across devices.
This allows training with long multimodal sequences while maintaining a sufficiently large global batch size.

A key systems bottleneck is the online preparation of heterogeneous media samples.
We therefore use an asynchronous server-based preprocessing pipeline for high-concurrency audio and video handling.
Rather than performing media I/O, audio-video extraction, and VAE encoding inside the training workers, dedicated data servers process samples asynchronously and prefetch ready-to-use training examples.
This reduces stalls from file access, media decoding, audio extraction, and feature preparation, improving GPU utilization during large-scale training.

To handle the heterogeneous sequence lengths of mixed-modality data, we use two batching strategies.
First, samples with similar sequence lengths are grouped into buckets to reduce padding overhead.
Second, each dataloader micro-batch contains samples from a single modality, which makes the per-step sequence length more predictable and reduces padding within each forward-backward pass.
We then interleave modalities across consecutive micro-batches through gradient accumulation, so that each effective global batch retains diverse multimodal supervision.
This improves efficiency of jointly training on audio, video and audio-video with substantially different token-length distributions.

Training is conducted in three stages.
Stages~1 and~2 use 160 NVIDIA H100 GPUs, together with asynchronous data servers, for approximately three weeks, corresponding to about $160 \times 21 \times 24 = 80{,}640$ H100 GPU-hours.
Stage~3 uses 160 NVIDIA H100 GPUs for one additional week, corresponding to about $160 \times 7 \times 24 = 26{,}880$ H100 GPU-hours.
Overall, the full training pipeline requires approximately $107{,}520$ H100 GPU-hours, assuming continuous training.

\end{document}